\documentclass[runningheads]{llncs}
\usepackage{graphicx}
\usepackage{booktabs}
\usepackage{multirow} 
\usepackage{hyperref}
\usepackage{amsmath}
\usepackage[dvipsnames]{xcolor}

\begin{document}
\title{Robust Curve Detection in Volumetric Medical Imaging via Attraction Field}
\titlerunning{Curve Detection}
% If the paper title is too long for the running head, you can set
% an abbreviated paper title here
%
%
\author{Farukh Yaushev\inst{1, 2} \and Daria Nogina\inst{1, 3} \and Valentin Samokhin\inst{1, 2} \and Mariya Dugova\inst{1} \and Ekaterina Petrash\inst{1} \and Dmitry Sevryukov\inst{1} \and Mikhail Belyaev\inst{4} \and Maxim Pisov\inst{4}}
\authorrunning{F. Yaushev et al.}
% First names are abbreviated in the running head.
% If there are more than two authors, 'et al.' is used.
%
\institute{
IRA-Labs Ltd, Moscow, Russia \and
Kharkevich Institute for Information Transmission Problems, Moscow, Russia \and
Lomonosov Moscow State University, Moscow, Russia \and
AUMI AI Limited, London, United Kingdom
\\
\email{f.yaushev@ira-labs.com}
}
\maketitle              % typeset the header of the contribution
\setcounter{footnote}{0}

\begin{abstract}
Understanding body part geometry is crucial for precise medical diagnostics. Curves effectively describe anatomical structures and are widely used in medical imaging applications related to cardiovascular, respiratory, and skeletal diseases. Traditional curve detection methods are often task-specific, relying heavily on domain-specific features, limiting their broader applicability. This paper introduces an novel approach for detecting non-branching curves, which does not require prior knowledge of the object's orientation, shape, or position. Our method uses neural networks to predict (1) an attraction field, which offers subpixel accuracy, and (2) a closeness map, which limits the region of interest and essentially eliminates outliers far from the desired curve. We tested our curve detector on several clinically relevant tasks with diverse morphologies and achieved impressive subpixel-level accuracy results that surpass existing methods, highlighting its versatility and robustness. Additionally, to support further advancements in this field, we provide our private annotations of aortic centerlines and masks, which can serve as a benchmark for future research. The dataset can be found at \url{https://github.com/neuro-ml/curve-detection}.

\end{abstract}
\section{Introduction}
\label{sec:intro}

Semantic segmentation is inarguably one of the most prominent tasks in medical image analysis with a particular focus on volumetric images such as computed tomography (CT) and magnetic resonance imaging (MRI)~\cite{minaee2021image}. Indeed, various applications such as brain tumor~\cite{bakas2018identifying}, ischemic stroke~\cite{winzeck2018isles}, or lung cancer~\cite{setio2017validation} detection and localization stimulated the development of many $3$D segmentation networks~\cite{resunet}. 

While many parts of the human body can be effectively represented as volumetric objects with corresponding segmentation masks, there are numerous objects with an underlying \textit{low-dimensional} structure that cannot be accurately described solely through segmentation masks. Important examples include the aortic centerline, vertebral column centerline, and ureters, which are essentially\textit{curves}. A comprehensive understanding of the geometry associated with these curves is crucial for precise disease detection and severity assessment~\cite{zhang2013robust,anchor-free-vertebrae,wang2021automatic,hadjiiski2014ureter,spencer1996image}.

Curve detection in medical images presents several significant \textit{challenges}. \textit{Firstly}, the complex geometry of the structures being sought poses difficulties because curves within the human body can exhibit diverse orientations, localization, and, notably, significant curvature. \textit{Secondly}, many curves in medical images are not bound to the edges of objects or organs. For example, the aortic centerline is defined by surrounding vessel semantics rather than explicit local signals solely bound to the centerline. This requires the detection method to infer the curve from surrounding anatomical and contextual information instead of relying solely on edge detection or segmentation. \textit{Finally}, most medical imaging problems require detecting only anatomically relevant structures, omitting other potential proposals. Therefore, the method should be capable of differentiating similar local structures by carefully weighing the global context.

Numerous approaches have been proposed for curve detection. The most straightforward one is a direct segmentation at the pixel level~\cite{zhang2018geometric,valente2017real}. However, the accuracy of this method is limited by the resolution of the raster, and it is not prone to the continuous nature of curves. Another class of approaches closely related to curve localization is edge and line segment detection methods, including classical algorithms like the Canny edge detector~\cite{canny-edge-detection} or LCD~\cite{von2008lsd}, and modern deep learning-based methods~\cite{attraction-field,deep-lsd,xie2015holistically,xu2021line,poma2020dense}. A notable example is~\cite{attraction-field}, where authors use a convolutional neural network (CNN) to predict an \textit{attraction field}: for each pixel, the displacement to the closest edge is predicted, which is then used to generate a point cloud of potential edges. \textit{Attraction field} proved to be a powerful concept, and several extensions have been proposed~\cite{deep-lsd,xue2020holistically}. Despite these methods' broad applicability, they currently face the above-mentioned \textit{challenges} in the medical domain, as we show for the relevant ones.

The curve detection also often appears in \textit{medical imaging} domain. However, unlike edge or line detection methods, approaches in this field frequently rely heavily on task-specific features, limiting their generalizability. For instance, some methods depend on a heuristic starting point or the image's orientation~\cite{zhao2021automatic,anchor-free-vertebrae,deep_reinforcement}. Other methods rely on segmenting objects that form the curve, like the aorta or blood vessels~\cite{aorta_skeleton,rouge2023cascaded,shit2021cldice,centerline-extraction1,centerline-extraction2}, which limits their use in tasks without an explicit object for segmentation, such as in the vertebral column centerline detection. Additionally, accuracy of segmentation- or heatmap-based methods~\cite{aorta_skeleton,centerline-extraction1} is limited by the raster’s resolution, making them overly sensitive to image spacing. These limitations highlight the need for more versatile and robust approaches in curve detection for medical applications.

In this article, we introduce a novel approach to detecting non-branching curves that surpasses current methods by combining their strengths while addressing their limitations. Our strategy offers a fresh perspective on solving the challenge of curve detection in medical imaging, adapting the concept of an \textit{attraction field} originally designed for line detection.

\textbf{Our contribution} is three-fold. \textit{First}, we introduce a new method for localizing non-branching curves in volumetric medical images, easily adaptable to various target objects. It is robust to the object's geometry and does not require task-specific knowledge of orientation, shape, or position. \textit{Second}, we demonstrate that the proposed method achieves subpixel-level accuracy and outperforms the existing approaches on two clinically significant and diverse tasks: aortic centerline and vertebral column centerline detection. \textit{Finally}, to enhance reproducibility and comparability, we release all our available annotations of aortic centerlines and masks on several open datasets. They include complex cases on which, as we show, conventional methods perform poorly. The dataset can be found at \url{https://github.com/neuro-ml/curve-detection}.
\section{Method}
\label{sec:method}

\begin{figure}[t]
   \includegraphics[width=\textwidth]{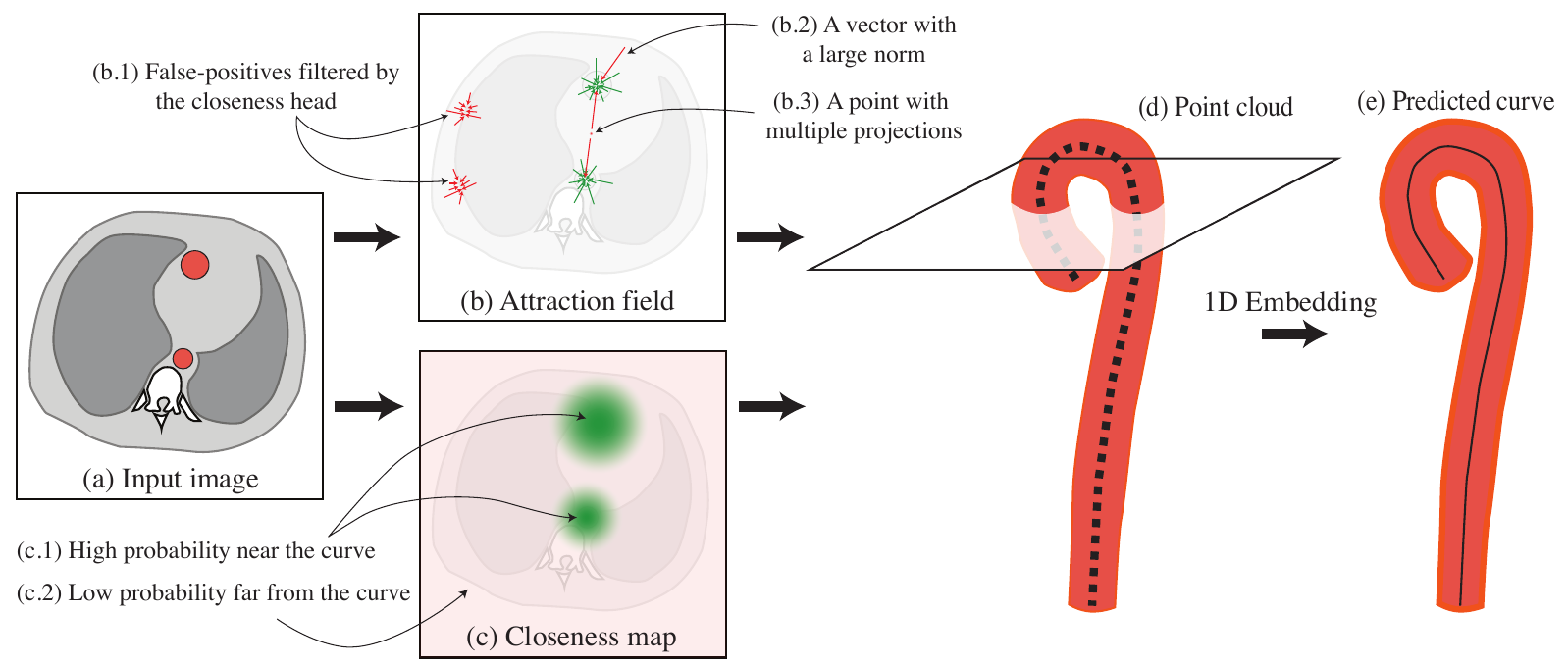}
    \caption{
        A schematic representation of the prediction pipeline for aortic centerline detection: a) a $3$D input image, red circles denote the aorta; b) the predicted attraction field depicted for a given $2$D slice, the red color indicates the field vectors that lead to inaccurate predictions; c) the closeness map predicted for a given $2$D slice; d) the same $2$D slice relative to the predicted point cloud, the black dots represent the unordered set of points; e) the result of ordering using Isomap~-- the final predicted $3$D curve.
    }
    \label{fig:inference}
\end{figure}

Our method Fig.~\ref{fig:inference} is based on a 3D two-headed CNN architecture with a VNet-based~\cite{resunet} backbone. The \textit{first} head predicts the attraction field, pointing to the closest location on the curve in each voxel Fig.~\ref{fig:inference}(b). The \textit{second} head predicts the closeness map for each voxel. It represents a region of interest around the curve Fig.~\ref{fig:inference}(c). At inference, predictions of two heads are combined. First, voxels are filtered according to the predicted closeness map and distances to the curve. Predicted attraction vectors of these voxels are then used to get the point cloud on the desired curve by shifting their coordinates. Finally, the point cloud is thinned using the non-maximum suppression algorithm, and then the points are parameterized using one-dimensional embedding.

\subsection{Loss Function}

The loss function consists of three pivotal components designed to refine specific aspects of the model's predictions.

\subsubsection{Attraction Field} The $L_{field}$, component addresses the loss related to the attraction field $F$~\cite{attraction-field}. For any given voxel $p$ within an image, its corresponding ground truth field vector, $F_p$, is calculated as the projection vector from $p$ to its nearest point $r_p$ on the ground truth curve: $F_p = r_p - p$. The norm of $F_p$ ($\|F_p\|_2$) is the distance to the curve. The network is trained to predict attraction field $\hat{F_p}$ Fig.~\ref{fig:inference}(c) for each voxel $p$ by optimizing a MSE loss: 

\begin{equation}
    L_{field}(\hat{F}, F) = \frac{1}{N} \sum_p \|\hat{F_p} - F_p\|\footnote{$\frac{1}{N} \sum_p$ denotes averaging over all voxels of the image}.
\end{equation}

\subsubsection{Closeness Map} In practice, CNNs' limited receptive fields~\cite{receptive-field} hinder their ability to capture long relations, leading to inaccurate attraction field predictions far from the sought structure. For this reason, we predict for each voxel a closeness $C_p = I[\|F_p\|_2 \leq R_c]$, where $I$ is the indicator function and $R_c$ is a hyperparameter that regulates the maximal distance at which the network is allowed to make predictions. Predicted closeness, $\hat{C}$, is utilized during inference to filter out voxels far from the desired curve. The closeness head Fig.~\ref{fig:inference}(c) is trained by optimizing the binary cross entropy: 

\begin{equation}
    L_{cls}(\hat{C}, C) = \frac{1}{N} \sum_p BCE(\hat{C}_p, C_p).
\end{equation}

\subsubsection{Handling multiple projections} Some voxels can have several projections on the curve Fig.~\ref{fig:inference}(b.3). This is a source of inconsistency that impedes the training of the regression head. Instead of picking a projection, the network averages all the possible directions, thus ending up with a near-zero prediction. Note that even though a point can have multiple projections, by definition, each projection will have the same norm, $\|F_p\|_2$. This observation motivates us to add a regularization term: 

\begin{equation}
    L_{norm}(\hat{F}, F) = \frac{1}{N} \sum_p \mid \|\hat{F_p}\|_2 - \|F_p\|_2 \mid.
\end{equation}

\subsubsection{Final Loss} The resulting loss is the sum of all the above components: 

\begin{equation}
L = L_{field} + L_{cls} + L_{norm},
\end{equation}

where the $L_{norm}, L_{field}$ are only computed inside the regions $\|F_p\|_2 \leq R_c$ and $\|F_p\|_2 \leq R_f$ respectively. $R_f$ should be chosen small enough to reduce the probability of encountering multiple projections problem.

\subsection{Inference}
\label{sec:method_inference}

The point cloud is obtained as follows: 
\begin{equation}
\{ \hat{F_p} + p \mid \hat{C_p} \geq t, \|\hat{F_p}\|_2 \leq R_f\}_p,
\end{equation}

where $p$ iterates over all the voxels on the image, $\hat{C_p}, \hat{F_p}, t, R_f$ are the predicted closeness, attraction field, and their respective thresholds. Fig.~\ref{fig:inference}(b,c) illustrates the motivation for both closeness- and norm-based filtration: the closeness head improves precision, and the regression head improves recall. 

To refine the predicted point cloud, we employ a classic non-maximum suppression technique~\cite{nms} where the \textit{closeness} function is defined by the Euclidean distance, and the \textit{confidence} is represented as $-\|\hat{F_p}\|_2$ Fig.~\ref{fig:inference}(d). After obtaining a point-reified cloud, we reorder the points to obtain a curve by reducing the dimensionality of the point cloud to $1$D using Isomap~\cite{isomap}. This effectively gives us the internal parametrization of the curve, which is used to order the points Fig.~\ref{fig:inference}(e).

\section{Experimental Setup}

\subsubsection{Data}

% Our method was evaluated on aortic centerline and vertebral column centerline detection tasks. Vertebral column annotations were obtained from \textbf{LungCancer500}~\cite{anchor-free-vertebrae} and \textbf{VerSe2020}~\cite{verse1}, while aortic centerlines were annotated in-house and shared on GitHub\footnote{\url{https://github.com/neuro-ml/curve-detection}}. Three experienced radiologists annotated each image, and their averaged annotations formed the ground truth curve.

Our method was evaluated on aortic centerline and vertebral column centerline detection tasks. Vertebral column annotations for $400$ images were obtained from \textbf{LungCancer500}~\cite{anchor-free-vertebrae} and \textbf{VerSe2020}~\cite{verse1}. Aortic centerlines for $142$ randomly selected images from \textbf{LIDC-IDRI}~\cite{lidc} and \textbf{AMOS}~\cite{amos} were annotated in-house and shared on GitHub\footnote{\url{https://github.com/neuro-ml/curve-detection}}. Three experienced radiologists annotated each image, and their averaged annotations formed the ground truth curve.

\subsubsection{Training Details}

We utilized a VNet-based backbone \cite{resunet} with two output heads for all experiments. Each head comprises six Residual blocks with a kernel size of $3$ and padding of $1$. Training employed $75,000$ iterations with Adam optimizer, batch size of $2$, and a learning rate policy starting at $10^{-4}$ and modified it to $5 \cdot 10^{-5}$ and $5\cdot 10^{-6}$ at iterations $22,500$ and $37,500$ respectively. Preprocessing included resampling to $2 \times 2 \times 2\: mm^3$ resolution, intensity normalization, and simple data augmentation: rotations, flips, transpositions, and random patch sampling of size $150 \times 150 \times 150$. In all our experiments, we use the following hyperparameters (see Section \ref{sec:method} for details): $R_c = 10, R_f = 5, t = 0.5.$ The dependence of the metric on changes in hyperparameters is detailed in the Table~\ref{tab:aorta-hyperparameter}.

\subsubsection{Metrics}

For all tasks, we use the $1$D variants of metrics between sets of points: Hausdorff Distance (\textbf{HD}), the Average Symmetric Surface Distance (\textbf{ASSD}), and the Surface Dice (\textbf{SD}) introduced in~\cite{surface-dice}, for the thresholds $1$ $mm$ and $3$ $mm$. All results are computed via $5$-fold cross-validation.

\subsubsection{Baselines}

To validate our method, we compare it with various baselines, drawing inspiration from both general methods and task-specific strong\footnote{\textit{Strong baseline} is a baseline that significantly exploits the unique features of a particular task, like the spine's vertical orientation.} baselines for curve detection.

\begin{enumerate}
    \item  \textbf{Skeleton}: The skeletonization-based approach is widely used for extracting tubular organ centerlines~\cite{grelard2017new,aorta_skeleton}. It involves segmenting the aorta followed by applying a skeletonization algorithm to extract its centerline. However, it is unsuitable for vertebral column centerline detection due to the absence of a specific tubular object for segmentation.
    
    \item \textbf{Soft-argmax}: A strong soft-argmax-based vertebral column centerline detection method, leveraging the vertical orientation of the backbone~\cite{anchor-free-vertebrae}. However, it is unsuitable for aortic centerline detection because it assumes each axial slice intersects the curve at most once, which does not hold for the complex geometry of the aorta, especially near its arch.
    
    \item \textbf{Seg}: This method involves straightforward curve segmentation using binary masks with $1$-voxel wide curves.
    
    \item \textbf{Htmp}: Heatmap-based techniques, frequently used in key-point detection tasks, have demonstrated promising results~\cite{heatmap-1,heatmap-2}. Therefore, we compared our method with such an approach, which predicts distances (\textit{heatmap}) to curve within the predicted closeness map. Points onto the curve are obtained by thresholding the \textit{heatmap}. This approach is close to ours but does not use displacement vectors; it only considers their norms.
    
    \item \textbf{Att}: We directly compare with the method proposed in~\cite{attraction-field}, which shows state-of-the-art results in the task of line segment detection. We trained the network to predict the vectors of the attraction field solely for each voxel. Unlike our method, it does not include the \textit{closeness} head.
    
\end{enumerate}

\begin{table}[t]
\caption{
    \label{tab:all-metrics}
    Comparison of curve detection methods' metrics and their standard deviation (in brackets) for the aortic centerline and vertebral column centerline detection tasks. Bold numbers indicate the best performance. Omitted cells refer to methods not applicable to the task.
}
\begin{tabular}{cc|c|c|c|c|c|c}
    \toprule 
    Task & \parbox{4em}{\centering Metrics} & Skeleton & Soft-argmax & Seg & Htmp & Att & Ours \\ 
    \midrule 
    & \multicolumn{7}{c}{AMOS, LIDC-IDRI} \\
    \cmidrule(lr){2-8}
    \multirow{4}{3em}{ \rotatebox[origin=c]{90}{ \parbox{5em}{ \centering Aortic \\ centerline}}} & SD-1 & 0.30(0.11) & - & 0.43(0.12) & 0.49(0.16) & 0.43(0.17) & \textbf{0.56(0.20)} \\
    & SD-3 & 0.88(0.16) & - & 0.91(0.11) & 0.95(0.05) & 0.93(0.12) & \textbf{0.97(0.04)} \\
    & HD & 21(21) & - & 22(27) & 17(25) & 21(28) & \textbf{15(16)} \\
    & ASSD & 2.5(2.2) & - & 2.5(3.0) & 2.1(4.5) & 2.9(5.4)& \textbf{1.4(1.1)} \\

    \midrule 
    \multirow{4}{2.5em}{\rotatebox[origin=c]{90}{ \parbox{20em}{ \centering Vertebral column \\ centerline}}} & \multicolumn{7}{c}{Cancer500} \\
    \cmidrule(lr){2-8}
     & SD-1 & - & 0.36(0.24) & 0.14(0.16) & 0.34(0.21) & 0.24(0.26) & \textbf{0.42(.30)} \\
    & SD-3 & - & 0.94(0.11)& 0.61(0.30) & 0.94(0.08) & 0.87(0.19) & \textbf{0.96(.06)} \\
    & HD & - & 8(3) & 31(33) & 6(2) & 11(10) & \textbf{4(2)} \\
    & ASSD & - & 1.6(0.7) & 7.6(15.1) & 1.5(0.5) & 2.3(1.3) & \textbf{1.3(0.5)} \\
    
    \cmidrule(lr){2-8}
    & \multicolumn{7}{c}{VerSe val} \\
    \cmidrule(lr){2-8}
    & SD-1 & - & 0.17(0.23) & 0.36(0.17) & 0.19(0.20) & 0.31(0.26) & \textbf{0.45(0.27)} \\
    & SD-3 & - & 0.80(0.23) & 0.90(0.09) & 0.80(0.23)	 & 0.89(0.12) & \textbf{0.93(0.01)} \\
    & HD & - & 28(12) & 35(32) & 28(11) & 43(97) & \textbf{24(14)} \\
    & ASSD & - & 2.8(1.2) & 3.5(4.1) & 2.7(1.1) & 5.3(22.5) & \textbf{1.9(1.2)} \\

    \cmidrule(lr){2-8}
    & \multicolumn{7}{c}{ VerSe test} \\
    \cmidrule(lr){2-8}
    & SD-1 & - & 0.21(0.28) & 0.32(0.18) & 0.19(0.19) & 0.27(0.28) & \textbf{0.45(0.30)} \\
    & SD-3 & - & 0.81(0.23) & 0.89(0.13) & 0.75(0.29) & 0.85(0.18) & \textbf{0.92(0.12)} \\
    & HD & - & 29(12) & 31(26) & 27(13) & 45(102) & \textbf{22(12)} \\
    & ASSD & - & 2.8(1.6) & 2.9(3.6) & 2.8(1.3) & 5.3(16.1) & \textbf{1.8(1.1)} \\
    \bottomrule
\end{tabular}
\end{table}

The same VNet-based backbone architecture and postprocessing procedure ~\ref{sec:method_inference} were used for all models.

% \parbox{5em}{\centering Soft\\argmax}

\section{Results}
\label{sec:results}
\begin{figure}
% \fbox{\rule{0pt}{2in} \rule{0.9\linewidth}{0pt}}
   \includegraphics[width=\textwidth]{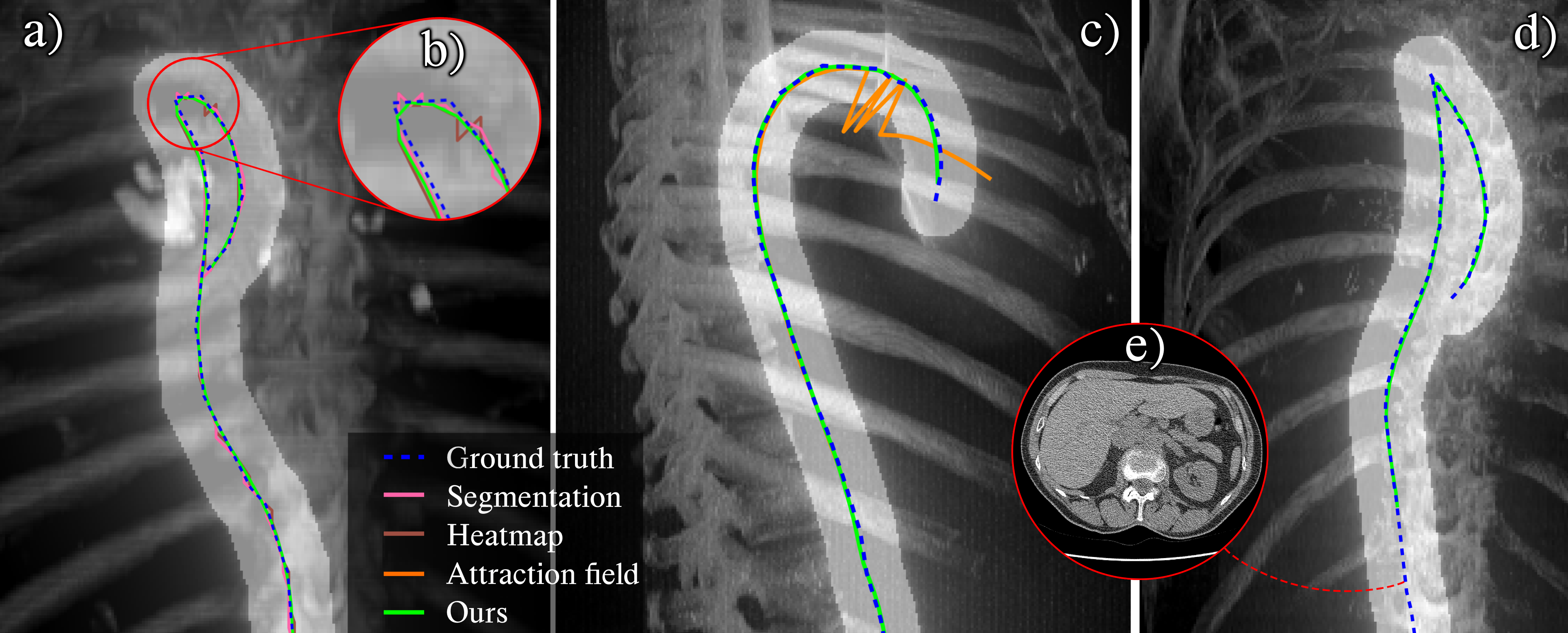}
    \caption{
    Aortic centerline predictions for various methods in sagittal or coronal projections, with the aorta highlighted in each case. Only relevant curves are displayed for clarity.
    a) Typical prediction without major defects;
    b) Magnified region showing the roughness of the \textit{Seg} model's predicted curve;
    c) Example of a false-positive for the \textit{Att} model caused by another tubular structure;
    d) Erroneous prediction by our method;
    e) Axial slice where our method did not make a prediction due to noisiness - the aorta is indistinguishable. 
    }
    \label{fig:aorta-predictions}
\end{figure}

\begin{figure}
\begin{center}
% \fbox{\rule{0pt}{2in} \rule{0.9\linewidth}{0pt}}
   \includegraphics[width=\textwidth]{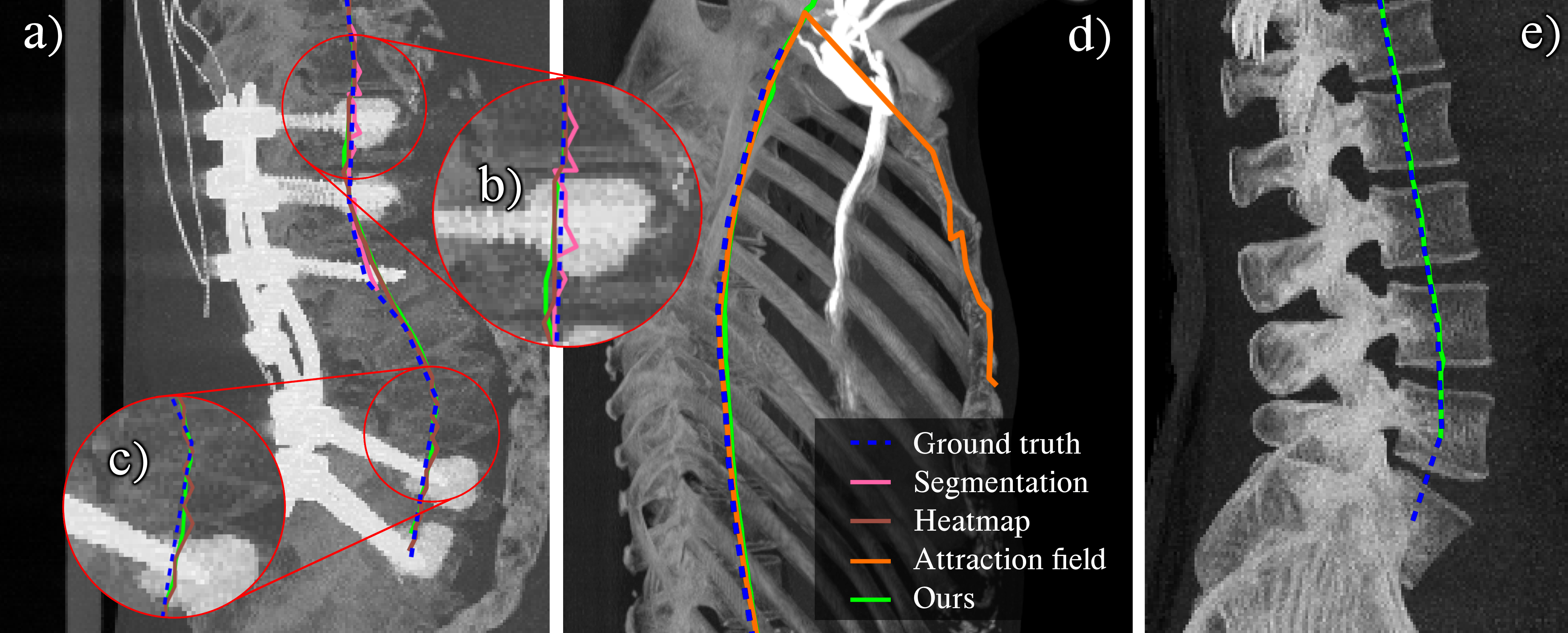}
\end{center}
    \caption{Vertebral column centerline predictions for several methods in sagittal projections. For clarity, only curves of interest are shown.
    a) a typical prediction without major errors from our method; 
    b) a magnified region highlighting the roughness of the predicted curve by the \textit{Seg} model; 
    c) a magnified region highlighting the roughness of the predicted curve by the \textit{Htmp} model;
    d) an example of a typical false-positive for the \textit{Att} model caused by another tubular structure; 
    e) an example of an erroneous prediction by our method.}
    \label{fig:backbone-predictions}
\end{figure}

Our model demonstrates consistently robust and accurate prediction ability, significantly outperforming baselines Table~\ref{tab:all-metrics}, Fig.~\ref{fig:aorta-predictions} and Fig.~\ref{fig:backbone-predictions}. Notably, our model surpasses baselines not only in terms of accuracy but in addressing specific challenges related to generability (\textit{Skeleton}, \textit{Soft-argmax}), prediction smoothness (\textit{Seg}, \textit{Htmp}), outliers far from the curve (\textit{Att}).

\subsection{Models Performance}

In this section, we analyze in detail and compare the performance of baselines and the proposed method separately for two tasks.

\subsubsection{Aortic Centerline}

Metrics for task-specific \textit{Skeleton} baseline are significantly inferior to \textit{Our} Table~\ref{tab:all-metrics}. This is primarily due to the skeletonization algorithm's dependency on the quality of the predicted mask, making it highly sensitive to minor errors within the segmentation.

Fig.~\ref{fig:aorta-predictions} demonstrates several typical predictions for \textit{Seg}, \textit{Htmp}, \textit{Att}, and \textit{Our} method. Note that the segmentation-based model yields sharp curves even in relatively simple cases because its accuracy is limited by the raster's resolution Fig.~\ref{fig:aorta-predictions}(b). The high HD value for \textit{Seg} is explained by false-negative predictions due to the highly imbalanced segmentation problem. The \textit{Htmp} model shows a similar behavior but with a lower occurrence of false-negative predictions, as evidenced by a lower HD value. On the other hand, the \textit{Att} model tends to produce false-positives around tubular structures that resemble the aorta, such as the pulmonary trunk Fig.~\ref{fig:aorta-predictions}(c). 

A rare issue observed with \textit{Our} method is occasional under-prediction of the curve's limits Fig.~\ref{fig:aorta-predictions}(d). However, each failure has an explicable reason: noisy image region Fig.~\ref{fig:aorta-predictions}(e) or scanning artifacts. 

\begin{figure}[t]
\begin{center}
% \fbox{\rule{0pt}{2in} \rule{0.9\linewidth}{0pt}}
   \includegraphics[width=0.5\linewidth]{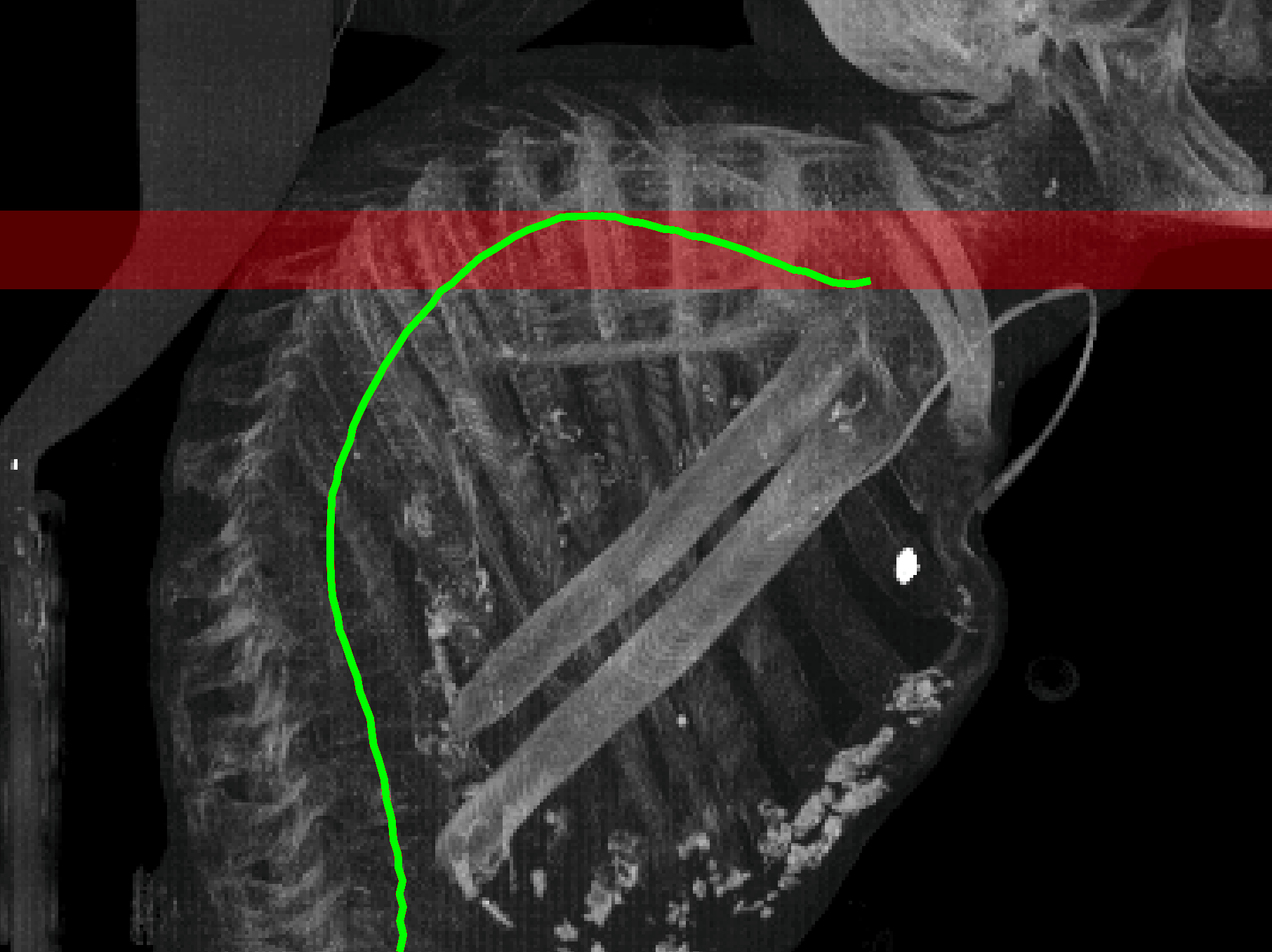}
\end{center}
    \caption{A sagittal projection of an image with kyphosis. The red region indicates the range of axial slices intersecting the curve at two locations~-- the cause of the inapplicability of soft-argmax-based methods. The green curve represents the prediction generated by our model.}
    \label{fig:curved-backbone}
\end{figure}

\subsubsection{Vertebral Column Centerline}

\textit{Our} approach outperforms the strong \textit{Soft-argmax} baseline on both the relatively simple \textbf{Cancer500} and more diverse \textbf{VerSe} dataset. Moreover, unlike \textit{Soft-argmax}, our method does not rely on additional structural knowledge of the vertebral column's shape, orientation, or position. This allows us to accurately localize the vertebral column centerline even in cases of extreme spinal curvature, such as severe kyphosis Fig.~\ref{fig:curved-backbone}.

The predictions made by other models exhibit behaviors similar to those seen in aortic centerline predictions Fig.~\ref{fig:backbone-predictions}. The \textit{Att} model, for instance, shows the same typical false-positive detections. Similarly, the \textit{Seg} and \textit{Htmp} baselines struggle to produce accurate and smooth curves due to their inherent limitations related to raster resolution constraints.

\subsubsection{Summary} 

Our model outperforms analyzed baselines in terms of accuracy and generality across both tasks. Overall, our method's observed Hausdorff Distance (HD) may appear high. This is because HD is generally very sensitive not only to false-positives (outliers) but also to false-negatives (e.g., near curve's limits), which our model is prone to Fig.~\ref{fig:aorta-predictions}(d). However, the more stable ASSD and Surface Dice metrics show that such errors are relatively rare. Furthermore, recall that our networks are trained with a $2 \times 2 \times 2\: mm^3$ spacing, which means that our method achieves \textit{subpixel-level} accuracy: ASSD $< 2\: mm$.

\begin{table}[t]
\caption{
    \label{tab:aorta-hyperparameter}
    The dependence of metric on the values of hyperparameters $R_c$ and $R_f$. Note that overall, changing the parameters in a reasonable range gives a quality still superior to the quality of baselines. However, for the aortic centerline, excessively large values of $R_c$ and $R_f$ can diminish performance due to limited receptive field and challenges related to multiple projections. This highlights the significance of optimizing the training loss component $F_{field}$ specifically near the curve.
}
\begin{center}
\begin{tabular}{c|cccc}

    \toprule 
    Hyperparameter & SD-1 mm & SD-3 mm & HD & ASSD \\

    \midrule 
    & \multicolumn{4}{c}{Aortic centerline, AMOS, LIDC-IDRI} \\
    \midrule
    $R_c = 5,$ $R_f = 5$ & 0.52 (0.14) & 0.95 (0.05) & 16 (18) & 1.7 (1.2)\\
    $R_c = 10,$ $R_f = 5$ & \textbf{0.56 (0.20)} & \textbf{0.97} (0.04) & \textbf{15 (16)} & \textbf{1.4 (1.1)} \\
    $R_c = 15,$ $R_f = 5$ & 0.54 (0.12) & 0.96 (0.05) & 16 (17) & 1.6 (1.3)\\
    $R_c = 20,$ $R_f = 5$ & 0.52 (0.21) & 0.97 (0.06) & \textbf{15 (16)} & 1.4 (1.3)\\
    $R_c = 10,$ $R_f = 10$ & 0.52 (0.18) & 0.94 (0.04) & 18 (15) & 1.8 (1.1)\\
    $R_c = 20,$ $R_f = 20$ & 0.50 (0.23) & 0.93 (0.06) & 18 (16) & 2.1 (1.4)\\

    \midrule 
    & \multicolumn{4}{c}{Vertebral column centerline, Cancer500} \\
    \midrule
    $R_c = 5,$ $R_f = 5$ & \textbf{0.44 (0.30)} & 0.97 (0.06) & 6 (12) & 1.3 (1.1) \\
    $R_c = 10,$ $R_f = 5$ & 0.42 (0.30) & 0.96 (0.06) & \textbf{4 (2)} & \textbf{1.3 (0.5)}\\
    $R_c = 15,$ $R_f = 5$ & 0.41 (0.30) & 0.97 (0.06) & 5 (2) & \textbf{1.3 (0.5)}\\
    $R_c = 20,$ $R_f = 5$ & 0.42 (0.30) & 0.97 (0.06) & 7 (12) & 1.4 (1.2)\\
    $R_c = 10,$ $R_f = 10$ & 0.40 (0.29) & 0.97 (0.06) & 7 (11) & 1.4 (1.0)\\
    $R_c = 20,$ $R_f = 20$ & 0.40 (0.30) & \textbf{0.97 (0.05)} & 5 (2) & \textbf{1.3 (0.5)}\\

    \bottomrule
\end{tabular}
\end{center}
\end{table}

\subsection{Ablation Study}
\label{sec:results-ablation}

For the ablation study, we demonstrate three aspects. (1) The importance of the \textit{closeness} head to filter out the outliers. \textit{Att} baseline produces many false-positives far from the ground truth curve. In contrast, such problems are not observed in our model, which highlights this head's importance in filtering out false-positives. (2) The importance of the \textit{attraction field} to achieve subpixel accuracy. \textit{Skeleton}, \textit{Seg}, and \textit{Htmp} baselines share a problem of prediction sharpness because of limited spacing. The displacement vectors provide the subpixel accuracy for our model. (3) The validity of the training strategy. We tested a reasonable range of hyperparameter values Table~\ref{tab:aorta-hyperparameter}. Within this range, our model's quality is still superior to the baselines'. However, for the aorta, overly large $R_c$ and $R_f$ values reduce the quality, probably due to a limited receptive field and multiple projection issues. This highlights the importance of optimizing the loss component $F_{field}$ near the ground truth curve.

% Minor $R_c$ variations had a negligible impact on the metrics. However, overly large $R_c$ values can cause outliers, as seen with the \textit{Att} model. Similarly, higher $R_f$ values worsen aortic centerline detection due to limited receptive field and multiple projection issues. This highlights the importance of training loss component $F_{field}$ near the ground truth curve.

\section{Discussion and Conclusion}

We have introduced and validated a novel method for detecting non-branching curves in medical imaging, surpassing existing approaches. To facilitate further advancements in this field, we have released our private annotations, which can serve as a benchmark for future research.

Our method \textit{directly} predicts coordinates of curve points, thereby preventing rounding errors that can occur in segmentation or heatmap-based methods. This allows our approach to fully utilize the information in the training data to construct smooth, subpixel-accurate curves.

To validate our method, we specifically focused on the tasks of detecting aortic and vertebral column centerlines. These tasks involve curves of \textit{different natures}. The aortic centerline is cane-shaped and cannot be parameterized using one of the coordinates. In contrast, the vertebral column centerline has a more straightforward shape but is not associated with any tubular structure. Despite this, our model consistently outperforms baselines in accuracy and generality across both tasks. 

It is important to note that our method is designed for non-branching curves and is not directly applicable to tasks involving anatomical configurations with branching structures, such as vascular networks or airway trees. However, our preliminary experiments suggest that our method can be extended to handle branching structures by incorporating the prediction of branching points.

{
\bibliographystyle{splncs04}
\bibliography{main}
}
\end{document}